\documentclass[conference]{IEEEtran}
\IEEEoverridecommandlockouts
\usepackage{cite}
\usepackage{amsmath}
\usepackage{flexisym}
\usepackage{breqn}
\usepackage{amssymb,amsfonts}
\usepackage{caption}
\usepackage{supertabular,booktabs}
\usepackage{algorithmic}
\usepackage{graphicx}
\usepackage{booktabs}
\usepackage{subcaption}
\usepackage{textcomp}
\usepackage{placeins}
\usepackage{longtable}
\usepackage{xcolor}
\usepackage{ulem}
\usepackage{url}
\usepackage[ frozencache,cachedir=.]{minted}

\def\BibTeX{{\rm B\kern-.05em{\sc i\kern-.025em b}\kern-.08em
    T\kern-.1667em\lower.7ex\hbox{E}\kern-.125emX}}
\begin{document}

\title{HiAER-Spike: Hardware-Software Co-Design for Large-Scale Reconfigurable Event-Driven Neuromorphic Computing
\thanks{}
}

 \author{
 
  \IEEEauthorblockN{Gwenevere Frank, Gopabandhu Hota, Keli Wang, Abhinav Uppal,  Omowuyi Olajide,\\
  Kenneth Yoshimoto, Leif Gibb, Qingbo Wang, Johannes Leugering, Stephen Deiss, and Gert Cauwenberghs}
  \IEEEauthorblockA{Institute for Neural Computation, UC San Diego, La Jolla CA 92093 \\
  {\small \{ghota, jfrank, k3wang, auppal, oolajide, kyoshimoto, lgibb, jleugering, sdeiss, gert\}@ucsd.edu, qingbo.wang@wdc.com 
 }}
 \vspace{-1cm}
}


\IEEEoverridecommandlockouts

\IEEEpubid{\makebox[\columnwidth]{979-8-3315-4127-9/24/\$31.00~\copyright2024 IEEE \hfill}
\hspace{\columnsep}\makebox[\columnwidth]{ }}

\maketitle

\IEEEpubidadjcol

\begin{abstract}
In this work, we present HiAER-Spike, a modular, reconfigurable, event-driven neuromorphic computing platform designed to execute large spiking neural networks with up to 160 million neurons and 40 billion synapses - roughly twice the neurons of a mouse brain at faster-than real-time. This system, which is currently under construction at the UC San Diego Supercomputing Center, comprises a co-designed hard- and software stack that is optimized for run-time massively parallel processing and hierarchical address-event routing (HiAER) of spikes while promoting memory-efficient network storage and execution.
Our architecture efficiently handles both sparse connectivity and sparse activity for robust and low-latency event-driven inference for both edge and cloud computing. A Python programming interface to HiAER-Spike, agnostic to hardware-level detail, shields the user from complexity in the configuration and execution of general spiking neural networks with virtually no constraints in topology. The system is made easily available over a web portal for use by the wider community.
In the following we provide an overview of the hard- and software stack, explain the underlying design principles, demonstrate some of the system's capabilities and solicit feedback from the broader neuromorphic community. 
\end{abstract}


\section{Introduction}
Spiking neural networks (SNNs) mirror the inherently event-driven way information is processed in the human brain by encoding it into the timing of \textit{spikes}. In the field of event-based sensing, particularly with dynamic vision sensors (DVS), this has proven to be a powerful \cite{deng2020rethinking} and highly energy efficient mode of computing. It seems likely that neuromorphic VLSI hardware could thus overcome some of the limitations of conventional von-Neumann computing architectures and provide similar power savings for other applications. But since there is still lacking hardware and software support for running large-scale SNN simulations, research has been limited to much smaller - and hence less capable - networks than those used in the field of Deep Learning.

To address this limitation, we have created a hierarchical address-event routing platform for spike-based processing (HiAER-Spike) for the explicit purpose of training and deploying SNNs at scale. As many aspects of SNNs like neuron models and learning rules are still areas of active research, our system leverages a reconfigurable FPGA based computing architecture that allows for full customization and continuous improvements. Under the umbrella of the NSF-supported Computer and Information Science and Engineering (CISE) Community Research Infrastructure program, we make this system available to you, the research community, and the general public.

At present, we primarily target researchers focused on neuromorphic systems and SNNs while consolidating the system's features. In the next phase, we aim for wider adoption by a diverse cross-section of users in the broader STEM research community, e.g. for large-scale brain simulations and the development of novel AI algorithms. To make the platform accessible to such a diverse audience, we created an intuitive and user-friendly open-source software interface that shields novice users from the challenges of operating and configuring such highly specialized neuromorphic hardware. Building on extensive existing network and storage infrastructure for user access and data sharing at the San Diego Supercomputer Center, the HiAER-Spike platform is hosted and maintained through the Neuroscience Gateway (NSG \cite{sivagnanam_neuroscience_2020}) Portal, which already serves over 1,100 registered users in the scientific community.

The insights and community feedback gathered from this system will inform the development of more targeted ultra-low power ASIC SNN accelerators in the final phase of the project.

In the remainder of this paper, we first give an overview of the hardware stack and how it accommodates large SNNs. Then we present the software interfaces that allow users to remotely access and utilize the system. Finally, we demonstrate how the system operates on an exemplary event-based vision use-case.

\begin{figure}
  \includegraphics[width=0.8\columnwidth]
  {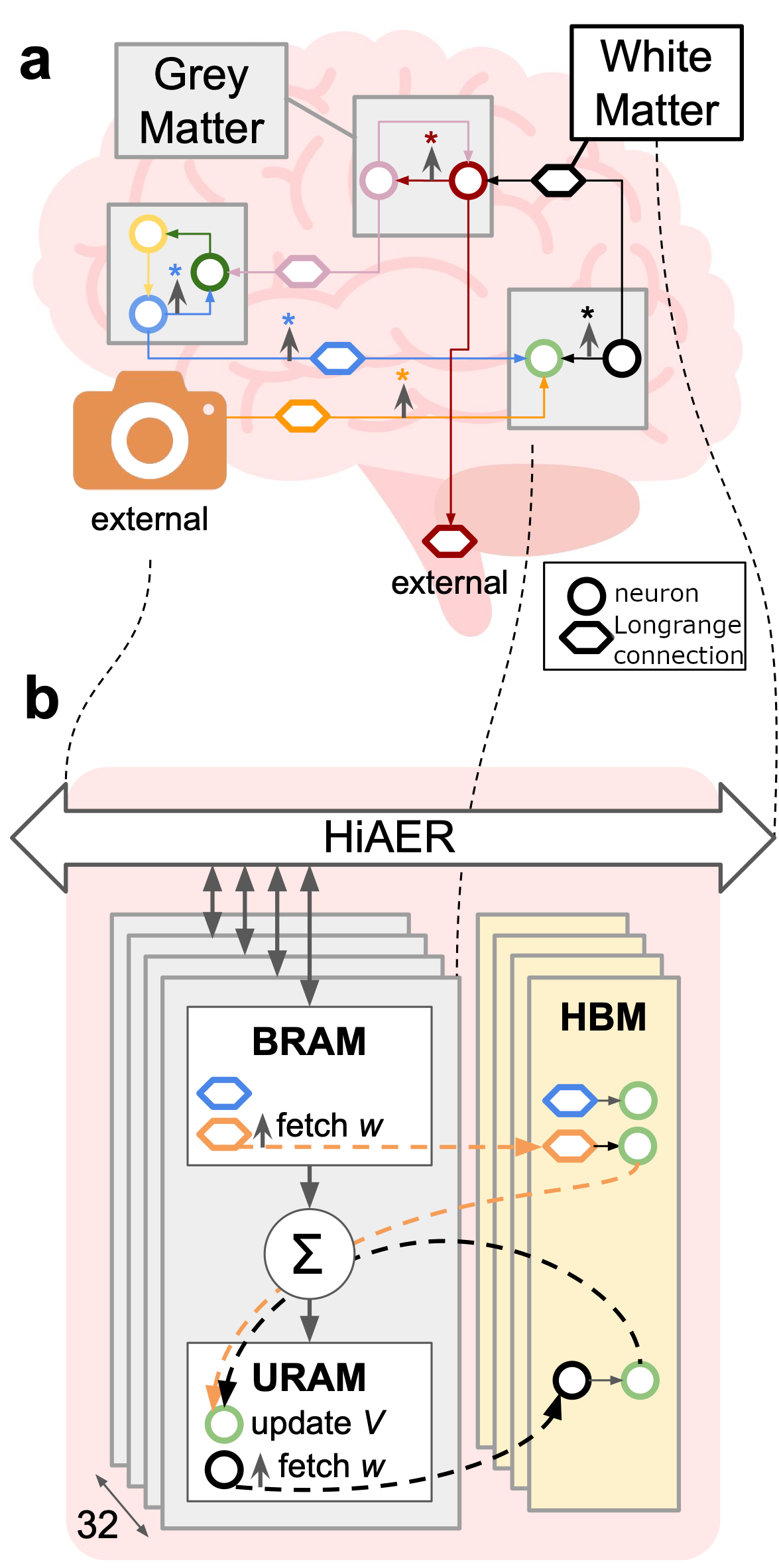}
  \caption{High-level system architecture of HiAER-Spike. (a) Neurons and synapses form the `grey matter' of dense local interconnects in the system while long-range connections (between cores and FPGAs) form the `white matter' of sparse global interconnects. (b) The hardware equivalent of (a) as implemented 
  in our multi-core architecture on the FPGA. The grey matter inside each core is implemented as sequentially updated integrate-and-fire neurons, whose internal state is stored in neural membrane registers in URAM, whereas spike events are routed through synaptic look-up tables in HBM. The white matter is implemented as a hierarchical multicast bus (HiAER) interconnecting axon spike register modules that are stored in BRAM across cores.
  }

  \label{fig:HiAER-Spike-top}
\end{figure}

\section{Related Works}
To the best of our knowledge, no FPGA-based solution of the scale we are proposing has ever been made publicly available to the community before. Inspired by the human brain, there have been efforts by multiple research groups to develop dedicated hardware for accelerating SNNs.
Perhaps most similar in scope is SpiNNaker \cite{painkras2012spinnaker}. SpiNNaker's architecture is structured around ARM-based chips designed by the APT Advanced Processor Technologies Research Group from the University of Manchester. The largest system contains 32400 chips and is capable of simulating hundreds of millions of neurons. The system is made available to researchers over a cloud interface. Although SpiNNaker is designed around custom chips that utilize a purpose-designed custom communication architecture the actual computations are performed using ARM processor cores as opposed to custom-designed logic.
Intel's neuromorphic platform Loihi 2 is designed around a custom architecture \cite{davies2021taking}. Loihi 2 chips consist of six processing units and 128 neuron cores per chip. Neuron cores allow for the specification of custom neuron models using a provided microcode. Loihi 2 supports up to a maximum of 1 million synapses per chip. Access to Loihi 2 devices is provided over a cloud interface to members of Intel's neuromorphic research community. BrainScaleS is a mixed-signal neuromorphic chip simulating 180,000 neurons and 40 million synapses. The upgraded platform, BrainScaleS-2 supports a more complex neuron and synapse model \cite{pehle2022brainscales }.
IBM recently introduced their NorthPole chip, an evolution of their True North design \cite{modha2023neural}. NorthPole is a custom architecture developed by IBM and supports up to 1048576 neurons per chip. NorthPole doesn't appear to be currently accessible to the public.
In comparison to existing works, HiAER-Spike is designed to leverage the advantage of having a widely available system based on reconfigurable hardware rather than fixed ASIC designs or more general microprocessor cores in order to rapidly incorporate user feedback on requested features into new revisions of the digital hardware designs in the form of new FPGA bitstreams and provide a community-driven testbed for iterating on new digital neuromorphic hardaware designs.

\section{Hardware System Organization and Software Co-design}
The architecture of HiAER-Spike utilizes a rack of 6 servers. Five of these servers are equipped with 8 high-end FPGA boards each featuring a Xilinx XCVU37P FPGA that includes 8GB of high bandwidth memory (HBM), and the sixth server is used as the control node to orchestrate the cluster and provide an interface to end users via the NSG Portal. Server CPUs coordinate with the FPGA processing cores via PCIe~3.0 to program network definitions and execute synaptic weight updates for various learning algorithms. 

Each FPGA, in turn, contains multiple SNN cores that run fully parallel with dedicated interfaces to the on-module SDRAM. Multiple layers of multicast address-event routing schemes (Network on Chip, FireFly, and Ethernet) \cite{park2016hierarchical, iscas_hiearahb} allow spikes to travel efficiently between cores, FPGAs, and servers.
Across all these levels of the hierarchy, the system tracks spike events with 1ms resolution and supports synaptic learning algorithms that require careful accounting for time differences between pre- and postsynaptic spikes, such as variations of spike timing-dependent plasticity (STDP).

This paper presents initial benchmarks using only a single operational core on one FPGA, and outlines how we intend to scale this up 1280-fold to multiple cores per FPGA across 40 FPGAs. We aim for each FPGA to support four million neurons and up to a billion synapses, yielding a total capacity of 160 million neurons and 40 billion synapses for the entire system - more than twice the number of neurons of a typical mouse brain\cite{herculano2006cellular} and at faster-than real-time simulation speed.

To be able to deploy networks at such scale, we have developed a network partitioning and resource allocation algorithm that assigns SNN simulation jobs to servers, FPGA boards, and cores as required to meet the user’s requirements \cite{mysore_hierarchical_2022}. Jobs can be submitted to NSG from anywhere in the world to run on HiAER-Spike hardware in the San Diego Supercomputer Center through the NSF ACCESS supercomputing network (NSF Program 21-555).

\section{Storing Networks}

\begin{figure}
  \includegraphics[scale=0.12] 
  {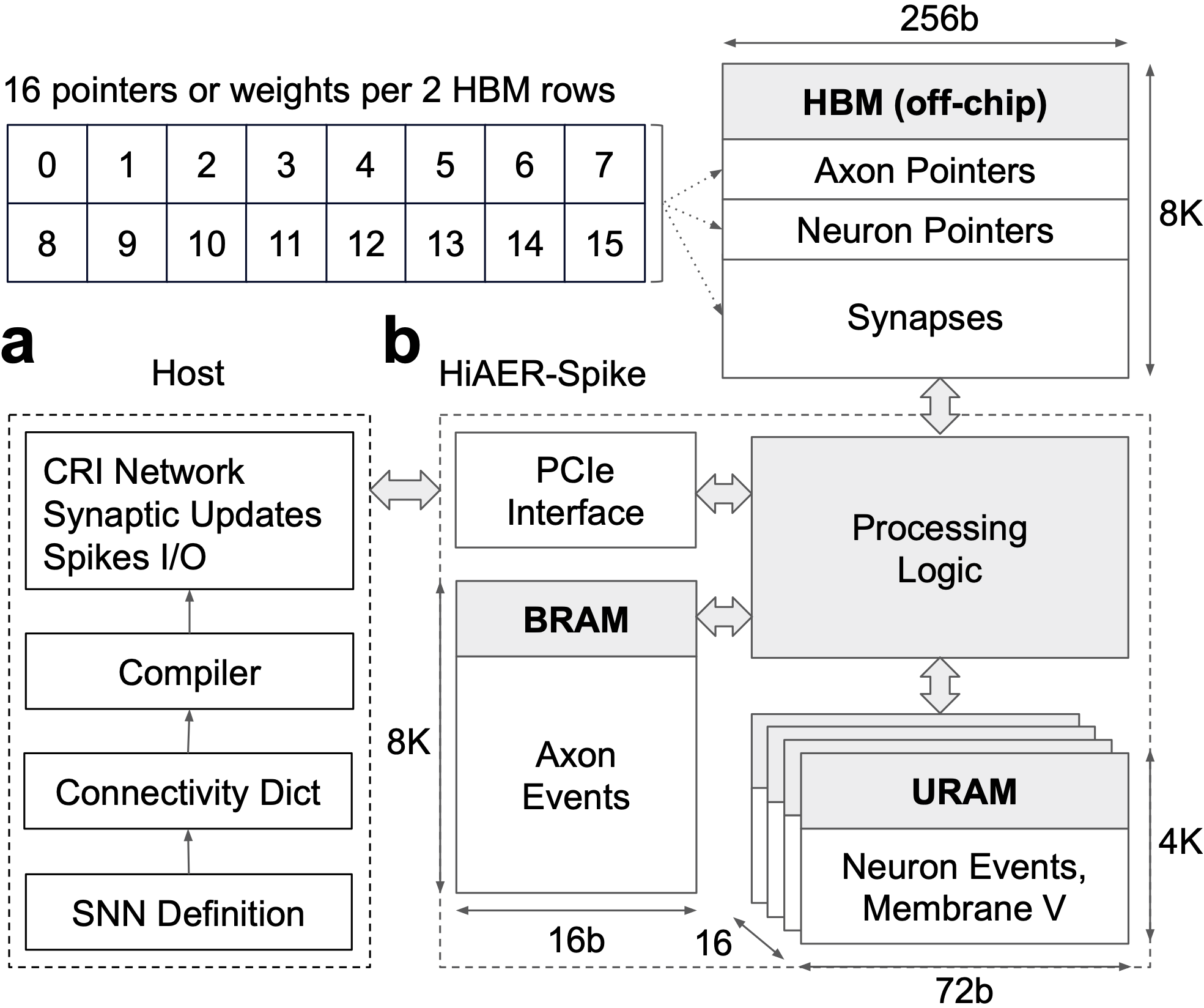}
  \caption{(a) Host programming interface with SNN compiler and low-level hardware interface. (b) Heterogeneous memory organization within a single core, as well as the off-chip HBM storing the synaptic connectivity table. On-chip URAM and BRAM store state variables of axons and neurons. The top-left panel shows the layout of the data structure in HBM, supporting parallelism of 16 neurons per single core.}

  \label{fig:process-flow}
\end{figure}

Fundamentally, the memory organization and data movement between compute units and memory determines the overall energy efficiency and latency of computation. For large-scale neuromorphic workloads, synaptic storage density is a key criterion in determining the efficiency and maximum network connectivity hardware can support. There have been various proposed data storage methodologies ranging from crossbar-based organization \cite{kulkarni2019neuromorphic,correll2020fully}, which is optimal for dense connectivity, to different variants of pointer-based organization, which offers higher storage efficiency in sparse networks \cite{pedroni2019memory}. Currently, deployed neuronal networks are often sparse, and several post-training optimizations such as quantization, and pruning further increase the sparsity. 
To support such workloads, we thus opted to store the connectivity table as adjacency lists rather than in a crossbar structure. The same pointer-based data structure is also suitable for storing the sparse activations of neurons. 

The network is stored inside the HBM in a format that includes pointers for neurons and axons (inputs to the network) that point to the respective synaptic weights. A portion of memory is reserved for axon and neuron pointers, and another portion is reserved for synapse definitions. Each pointer consists of a starting address and a number of rows in HBM that defines the region in memory where the outgoing synapses from the pointer's corresponding neuron or axon are defined.
The HBM with 8GB capacity per FPGA card is divided into segments of 16 slots spanning two rows (Fig. \ref{fig:process-flow}) with each slot storing a single pointer or synapse value. The network compiler is made aware of the memory alignment constraints of HBM, that is that synapses must utilize the same slot number as the pointer corresponding with their postsynaptic neuron, and adjusts the neuron and axon assignments to obtain maximum packing density in HBM, lowering execution latency. Having the pre-synaptic neuron pointer store just the base (start) address and the number of rows of HBM occupied by the post-synaptic connections, as opposed to absolute addresses further reduces memory usage. Neuron pointers are grouped by their corresponding neuron model in memory.

The routing of spikes then proceeds in two phases: first, for each neuron that fired in the previous time-step and for each incoming externally driven axon, the pointers to all post-synaptic connections are read into a queue. In the second phase, post-synaptic neuron addresses and the number of rows each neuron/axon's synapses span are retrieved for the enqueued pointers, corresponding synaptic weights are fetched from HBM, and the membrane potentials of the post-synaptic neurons are updated - possibly generating new spikes for the next cycle, and so on.

HBM allows accessing large packets over multiple ports, which we leverage to parallelize event lookups in the first phase and membrane potential updates in the second phase.
We use local on-chip Block-RAM (BRAM) and Ultra-RAM (URAM), laid out in the same structure as the HBM, to store spike events and membrane potentials, because they are queried every time step and thus have a dense access pattern. This hybrid approach, combining HBM and on-chip SRAM, provides significant energy and latency improvements. Further details of the hardware microarchitecture will be discussed in a future manuscript.

\section{Software Description}

We provide a software interface to allow end users to easily utilize the hardware. The hardware interface is implemented in C++ and Python and integrates with a community-developed Python library named L2S (for \textit{Lifelong Learning at Scale}) for defining and running networks on the hardware, which emerged out of a workgroup topic at the 2022 Telluride Neuromorphic Cognition Engineering Workshop (\url{https://sites.google.com/view/telluride-2022/home}).

\subsection{Supported neuron models}
\label{nmodels}
HiAER-Spike supports networks composed of multiple types of neurons. Currently simple binary neurons, that is neurons that either spike or don't at each timestep and accumulate no membrane potential between steps, and leaky integrate-and-fire neurons with an optional addition of noise to the membrane potential at each timestep are supported. Networks may be composed of multiple types of stochastic LIF and binary neurons with different parameters. ANN neurons have a single threshold parameter and each LIF neuron has three parameters: threshold, shift, and leak. Threshold determines the membrane potential at which the neuron spikes and the potential is reset to zero. Shift controls the magnitude of random noise added to the membrane potential at each step. Noise is a 17 bit signed integer randomly generated and then right shifted by the shift value if shift is negative or left shifted by the shift value if shift is positive. Finally leak controls the voltage leakage that occurs during each timestep $voltage=voltage-voltage/2^{leak}$.

\subsection{API Definition}
\begin{figure}
  \includegraphics[scale=0.09]{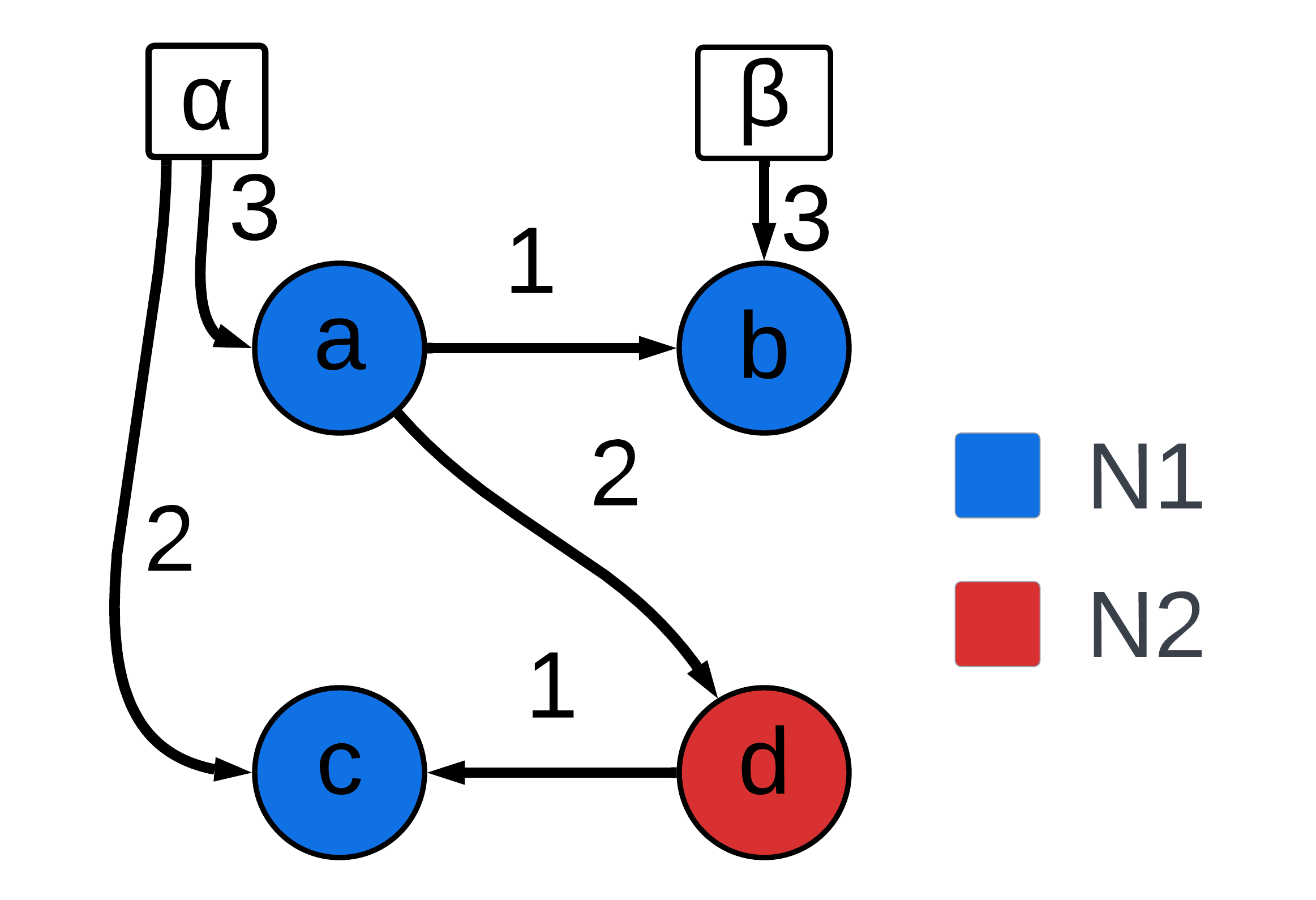}
    \begin{minted}[
frame=lines,
framesep=2mm,
baselinestretch=1.2,
fontsize=\footnotesize
]{python}
from l2s.api import CRI_network

N1 = LIF_neuron(threshold = 3, shift = -17, leak = (2**6)-1)
N2 = LIF_neuron(threshold = 5, shift = -17, leak = 1)
    
axons = {'alpha': [('a', 3),('c', 2)],
             'beta': [('b', 3)]}
             
neurons = {'a': (N1, [('b', 1), ('d', 2)]),
                   'b': (N1, []),
                   'c': (N1, []),
                   'd': (N2, [('c', 1)])}
                   
outputs = ['a', 'b']

network = CRI_network(axons=axons,
    neurons=neurons,
    outputs=outputs)
inputs = ['alpha','beta']
currSpikes = network.step(inputs)
  \end{minted}
  \caption{Example network and generating code}

  \label{fig:examp-network}

\end{figure}

The software exposes an API that allows users to define, run and interact with SNNs on the hardware. Networks can be defined by using a collection of simple Python objects. This section demonstrates how to create the network shown in Fig. \ref{fig:examp-network}, which consists of four neurons, $a$ through $d$, and two axons, $\alpha$ and $\beta$. In the example network neurons $a,b,$ and $c$ are leaky-integrate-and-fire neurons with no membrane potential noise and no leak (achieved by setting a large leak value) with a threshold of 3 and neuron $d$ is a leaky-integrate-and-fire neuron with a leakage parameter of 1 and and a threshold of 5 with no membrane potential noise.

First, the user must define a set of  neuron models:
\begin{minted}{python}
N1 = LIF_neuron(threshold = 3,
    shift = -17, leak = (2**6)-1)
N2 = LIF_neuron(threshold = 5,
    shift = -17, leak = 1)
\end{minted}

Each neuron is an instance of the \mintinline{python}{LIF_neuron} class. The \mintinline{python}{threshold}, \mintinline{python}{shift}, and \mintinline{python}{leak} variable affect the neuron model as described in section \ref{nmodels}.

Next, the user must supply an \mintinline{python}{axons} dictionary:
\begin{minted}{python}
axons = {'alpha': [('a', 3),('c', 2)],
             'beta': [('b', 3)]}
\end{minted}

Axons represent user controllable input from the external world into the network through which a user may send a spike to interact with multiple postsynaptic neurons. The keys in this dictionary are unique Python objects, usually strings, representing each axon coming into the network. The values associated with these keys are lists of tuples specifying the respective synaptic connections to one or more neurons in the network. Each tuple contains two elements, the postsynaptic neuron's unique key and an integer representing the weight of the connection. In this case we create the \mintinline{python}{"alpha"} axon with synapses to the \mintinline{python}{"a"} and \mintinline{python}{"c"} neuron with weights of \mintinline{python}{3} and \mintinline{python}{2}, respectively, and the \mintinline{python}{"beta"} neuron with a synapse to the \mintinline{python}{"b"} neuron with a weight of \mintinline{python}{3}.

The user must also supply a \mintinline{python}{neurons} dictionary:
\begin{minted}{python}
neurons = {'a': (N1, [('b', 1), 
    ('d', 2)]),
                   ('b': N2, []),
                   ('c': N2, []),
                   'd': (N2, [('c', 1)])}
\end{minted}

The keys in this dictionary are unique objects that represent each neuron in the network, usually a number or string. The corresponding values are tuples where the first element is a lists of all of the neurons' outgoing synapses and the second element is a neuron model used to specify membrane updates for the neurons. Each outgoing synapse is represented by a tuple comprising the postsynaptic neuron's unique key, and the integer weight of the synapse. In this case we instantiate the \mintinline{python}{"a"} neuron with synapses to the \mintinline{python}{"b"} and \mintinline{python}{"d"} neuron with weights of \mintinline{python}{1} and \mintinline{python}{2} respectively, neurons \mintinline{python}{"b"} and \mintinline{python}{"c"} with no outgoing synaptic connections, and the \mintinline{python}{"d"} neuron with synapses to neuron \mintinline{python}{"c"} with a weight of \mintinline{python}{1}.

Finally, the user must supply an outputs list:
\begin{minted}{python}
outputs = ['a', 'b']
\end{minted}

This is a list of all the neurons' unique keys whose spiking activity the user wishes to monitor. In this case we designate neurons \mintinline{python}{"a"} and \mintinline{python}{"b"} as output neurons.

Once the necessary data structures are defined, they may be passed to the constructor to create an instance of the 
\mintinline{python}{CRI_network} class that exposes functions through which the user can interact with the network:

\begin{minted}{python}
network =CRI_network(
    axons=axons,
    neurons=neurons,
    config=config,outputs=outputs)
inputs = ['alpha','beta']
currSpikes = network.step(inputs)
\end{minted}

The \mintinline{python}{step} method can be used to run one timestep of the network. The user supplies an inputs list consisting of the unique keys of each axon the user wishes to activate during the timestep. The function returns a list of all output neurons that spiked during the timestep. If the optional \mintinline{python}{membranePotential} flag is set, the function also returns the membrane potential for every neuron in the network. In this case both axons \mintinline{python}{"alpha"} and \mintinline{python}{"beta"} are activated.

The methods \mintinline{python}{read_synapse} and \mintinline{python}{write_synapse} can be used to change synapse weights after calling the 
\mintinline{python}{CRI_network} constructor. The \mintinline{python}{read_synapse} method takes the unique key for the presynaptic axon or neuron and the postsynaptic neuron and returns the corresponding synapse weight. The \mintinline{python}{write_synapse} method, accordingly, takes two keys and a new weight to set for the synapse. Here, we just increment the weight of the synapse from neuron \mintinline{python}{"a"} to \mintinline{python}{"b"} by one:
\begin{minted}{python}
currWeight = network.read_synapse('a', 'b')
network.write_synapse('a', 'b',
    currWeight+1)
\end{minted}

\subsection{Running Inference}
The L2S library enables users to define and interact with networks either in software simulation on their local machine, or accelerated on the HiAER-Spike hardware through NSG. The API of the L2S library remains exactly the same in both cases. If the user is running on their local machine where the HiAER-Spike hardware is not detected, network inference is run using a simulation of the hardware operations implemented in Python. If the user is running their code on the HiAER-Spike cluster over NSG, the L2S library detects the presence of the HiAER-Spike hardware and runs the inference on the accelerator hardware. This makes a seamless transition possible for users who want to develop on their local devices and then submit larger workloads to run on the HiAER-Spike cluster.

The simulator currently implements inference using  sparse matrix operations and fixed bit integer arithmetic. The network is represented by two sparse matrices holding the weights for axons and neurons, respectively. During the simulation, the membrane potentials are updated according to the chosen type of neuron. Then two binary vectors are constructed, holding the indices of all axons or neurons, respectively, that yielded a spike in that timestep. These vectors are multiplied by the axon and neuron weight matrices, respectively, to calculate the total input for each neuron in the network, and the membrane potentials are updated accordingly. Finally, an output spike is recorded at the current timestep for any output neuron that just fired.
\subsection{Managing HBM}\label{managing-hbm}

\begin{figure}
  \includegraphics[scale=0.55]{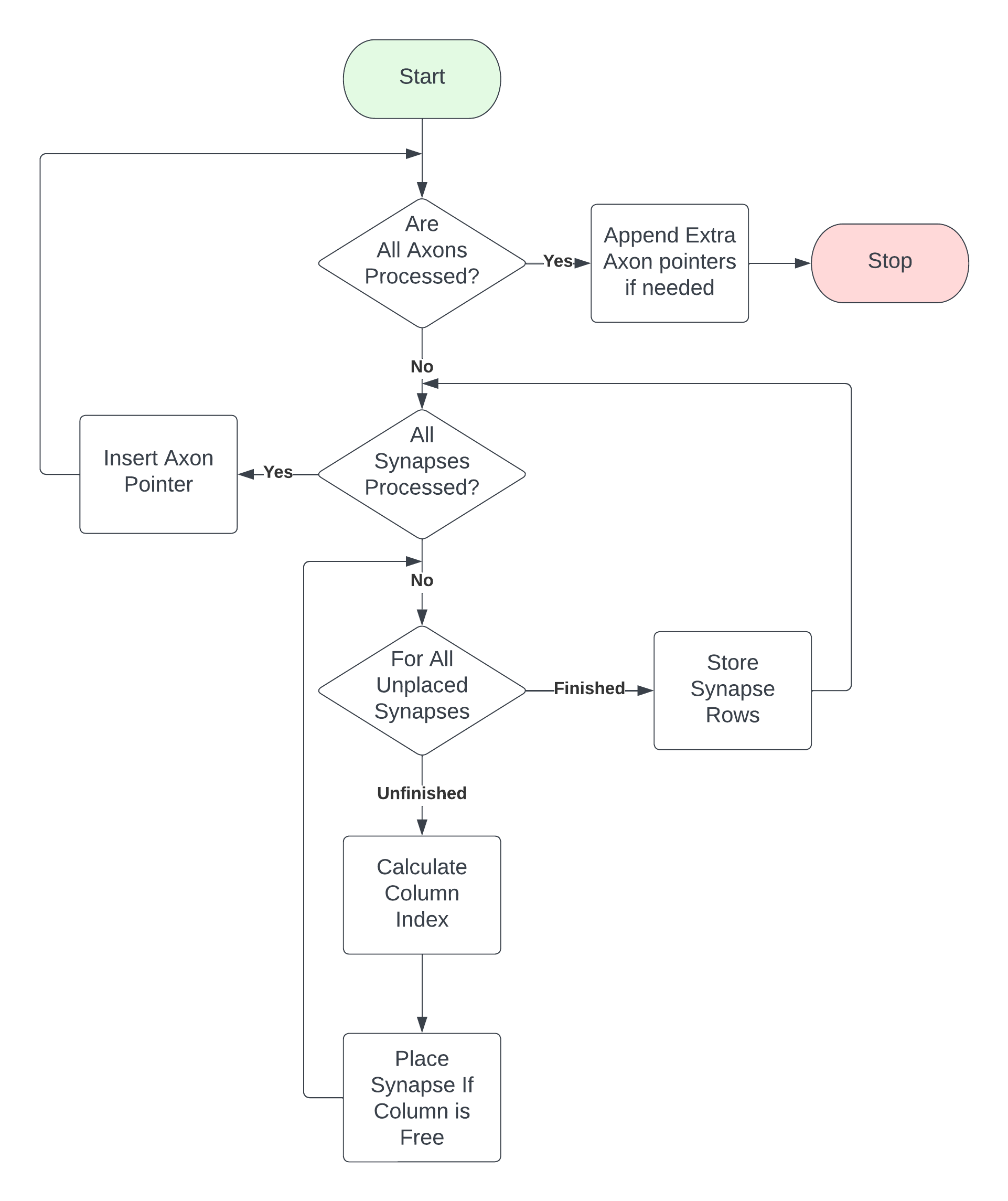}
  \caption{Simplified flowchart describing the process of mapping a network into HBM.}
  \label{fig:axon-flow}
\end{figure}

When running on the HiAER-Spike cluster, the software package must not only send commands to the FPGAs over PCIe to edit synapses, retrieve spike data, and orchestrate execution, but it must also initialize the network topology in HBM prior to execution. When a user creates a network, the software first partitions the network across all available cores/FPGAs/servers (if any) following the scheme described in \cite{mysore_hierarchical_2022}. After partitioning, the software maps the partitioned network into HBM for each core. Memory in HBM is divided into three sections, a section to hold axon pointers, a section to hold neuron pointers, and a section to hold synapses.

Axons are programmed into HBM according to the process described in (Fig.~\ref{fig:axon-flow}), which iterates through all axons and iterates through all synapses of each axon and  assign each a space in HBM.
All the synapses outgoing from an axon must be placed in a contiguous space in HBM. However synapse definitions in HBM must be aligned with the address of their post synaptic neuron. So each synapse for an axon is placed into memory following the needed alignment and once all the synapse definitions for a given axon are written into HBM an axon pointer is inserted into the axon pointer region of memory that points to the addresses in which the synapse definitions for that axon are stored.

Neurons are programmed into HBM in a similar manner, with two extra steps: First, in order to designate a neuron as an output neuron, a special flag must be set in the synapse definitions for that neuron. If necessary, the region in memory pointed to by a neuron can be expanded to accommodate this flag, by adding dummy synapses. Second if a neuron has no outgoing synapses an set of 16 zero weight synapses are inserted into HBM so that every neuron has a space in the section of HBM that holds synapses.

\section{Results}

Our hardware architecture and software pipeline is optimized for event-driven operation with different biologically inspired neurons. Here we present preliminary results for both energy consumption and latency that we observed for an exemplary event-based vision use-case, using only a single core of one FPGA card.

We trained a small convolutional spiking neural network on the IBM DVSGesture dataset \cite{amir2017low}, a dataset containing clips of 11 different gestures under 3 different lighting conditions recorded using a dynamic vision sensor camera. 

The network consists of 3 20-channel 2D convolutional layers each followed by a batch norm and then a 4500 to 512 fully connected linear layer and a 512 to 11 fully connected linear layer.

The network was first trained in SpikingJelly. Then the batch normalization layers were fused into the unrolled convolution layers, the network weights were quantized to integer values, and the resulting network was converted into the format used by HiAER-Spike and deployed on the hardware. The network as deployed comprises a total 33351 axons and 103680 neurons occupying approximately 83\% of a single core system and thus approximately .06\% of the total system capacity.

Table \ref{dvstable} shows the results of the network. An accuracy of 47.6\% was achieved with an average latency of 311 microseconds and an energy usage of 130 microjoules per frame of presented data averaged across 2112 frames in the testing dataset. The hardware's energy usage is primarily dominated by HBM accesses, thus energy consumption was approximated by the product of the energy cost of a single HBM access and the number of HBM accesses performed during a timestep.

\begin{table}[!t]
\renewcommand{\arraystretch}{1.3}
\caption{DVS Gesture Results}
\label{dvstable}
\centering
\begin{tabular}{|l|l|l|l|}
\hline
\# Neurons & Accuracy & Latency & Energy (approx.) \\ \hline
103680     & $47.6\%$   & $311\mu s$  & $130\mu J$ \\ \hline
\end{tabular}
\end{table}

\section{Conclusions}

We presented a large-scale FPGA-based high-throughput neuromorphic SNN accelerator platform designed to serve as a shared resource for the neuromorphic computing research community specifically, as well as STEM researchers more generally.
We described the hardware stack as well as the software interface that allow users to configure and run a broad class of SNNs remotely on the system.  Users are invited to submit jobs 
over the NSG portal in the form of a simple Python script, and encouraged to provide feedback and request (or add) new features for the ongoing development of the system.

First experiments on the HiAER-Spike show proof of concept that a single core of the system can run a relatively large spiking neural network for gesture recognition using DVS cameras with low latency and power consumption.

Future work will extend the HiAER-Spike platform in several directions. On the one hand, we intend to make the system more versatile, introducing further properties of interest for computational neuroscience modeling, such as more sophisticated neuron models and learning rules. On the other, we strive to improve performance and efficiency for more machine-learning oriented applications such as on-line adaptive pattern recognition. 
We believe that this approach can bring together both computational neuroscience and artificial intelligence communities that have traditionally pursued disparate computational approaches.

\section*{Acknowledgments}

This work has been supported by National Science Foundation CNS-1823366 (CRI: CI-NEW: Trainable Reconfigurable Development Platform for Large-Scale Neuromorphic Cognitive Computing), Office of Naval Research N00014-20-1-2405 (Science of AI Brain Inspired Next Generation Deep Learning: Efficient and Persistent Online Learning with Spikes), and Western Digital Corporation.  We thank Sankar Basu, Mitchell Fream, Kameron Gano, Justin Kinney, Duygu Kuzum, Tim Liu, Martin Lueker-Boden, Amit Majumdar, Justin Mauger, Thomas McKenna, Nishant Mysore, Emre Neftci, Bruno Pedroni, Terrence Sejnowski, Dejan Vucini\'c, Riley Zeller-Townson, and organizers and participants of the Telluride Workshop on Neuromorphic Cognition Engineering for key and insightful contributions in the development and application of the HiAER-Spike CRI. We also thank Tomas Whitlock, Kevin Roth, and Alexandros Kapouranis at Alpha Data Inc. for extensive FPGA applications advice, to Roger Miller at Samtec Corp. for FireFly help, Andrew Nelson at EXXACT Corp. for server help, Steven King at Arista for networking advice and Arista for donated switch hardware, Thomas Hutton and Christopher Cox at UCSD for switch configuration, Thomas Tate and all operations support staff at San Diego Supercomputer Center (SDSC) for installation help, Robert Buffington for IT support, Christopher Hughes of SDSC's High-Performance Computing group for networking support, and Andrew Ferbert, Ryan Nakashima and William Homan of SDSC's Research Data Services (RDS) division for server support and admin.

\FloatBarrier

\bibliographystyle{IEEEtran}
\bibliography{IEEEabrv,refs}

\end{document}